\documentclass[letterpaper]{article} 

\usepackage{aaai24}  
\def\copyright@on{}

\usepackage{times}  
\usepackage{helvet}  
\usepackage{courier}  
\usepackage[hyphens]{url}  
\usepackage{graphicx} 
\usepackage{natbib}  
\usepackage{caption} 
\usepackage{algorithm}
\usepackage{algorithmic}

\usepackage{newfloat}
\usepackage{listings}
\DeclareCaptionStyle{ruled}{labelfont=normalfont,labelsep=colon,strut=off} 
\floatstyle{ruled}
\newfloat{listing}{tb}{lst}{}
\floatname{listing}{Listing}
\newcommand{\para}[1]{{\vspace{2pt} \bf \noindent #1 \hspace{0.5pt}}}
\newcommand{\exbox}[2]{
{\begin{samepage}
\noindent
#1
\vspace{0.1cm}

\nopagebreak
\noindent
\fbox{
\begin{minipage}{0.45\textwidth}{
\begin{flushleft}
\footnotesize
\texttt{\noindent #2
}
\end{flushleft}}
\end{minipage}}
\end{samepage}
}}

\usepackage{enumitem}
\usepackage{multirow}
\usepackage{tabularray}
\usepackage{diagbox}
\usepackage{bbold}
\usepackage{colortbl}
\usepackage{makecell}
\usepackage{subfigure}
\usepackage{amsmath}
\usepackage{amssymb}
\usepackage{pifont}
\usepackage{amsthm}
\theoremstyle{definition}

\usepackage{amsfonts} 
\usepackage{MnSymbol}
\usepackage{wasysym}
\usepackage{tablefootnote}
\usepackage{threeparttable}
\usepackage{graphicx}
\usepackage{hhline}
\usepackage{booktabs}
\usepackage{siunitx}
\definecolor{mygray}{rgb}{0.784,0.784,0.784}

\title{MIA-Tuner: Adapting Large Language Models as Pre-training Text Detector}
\author{
    Wenjie Fu$^{1}$
    Huandong Wang$^{2}$
    Chen Gao$^{2}$
    Guanghua Liu$^{1}$
    Yong Li$^{2}$
    Tao Jiang$^{1}$
}
\affiliations{
    $^1$Huazhong University of Science and Technology\\
    $^2$Tsinghua University\\

    wjfu99@outlook.com
}

\usepackage{bibentry}

\begin{document}

\maketitle

\begin{abstract}
The increasing parameters and expansive dataset of large language models (LLMs) highlight the urgent demand for a technical solution to audit the underlying privacy risks and copyright issues associated with LLMs. Existing studies have partially addressed this need through an exploration of the pre-training data detection problem, which is an instance of a membership inference attack (MIA). This problem involves determining whether a given piece of text has been used during the pre-training phase of the target LLM. 
Although existing methods have designed various sophisticated MIA score functions to achieve considerable detection performance in pre-trained LLMs, how to achieve high-confidence detection and how to perform MIA on aligned LLMs remain challenging. 
In this paper, we propose MIA-Tuner, a novel instruction-based MIA method, which instructs LLMs themselves to serve as a more precise pre-training data detector internally, rather than design an external MIA score function. Furthermore, we design two instruction-based safeguards to respectively mitigate the privacy risks brought by the existing methods and MIA-Tuner.
To comprehensively evaluate the most recent state-of-the-art LLMs, we collect a more up-to-date MIA benchmark dataset, named WIKIMIA-24, to replace the widely adopted benchmark WIKIMIA. We conduct extensive experiments across various aligned and unaligned LLMs over the two benchmark datasets. The results demonstrate that MIA-Tuner increases the AUC of MIAs from 0.7 to a significantly high level of 0.9. Our code and dataset are available in the following link\footnote{\url{https://github.com/wjfu99/MIA-Tuner}.}

\end{abstract}

\section{Introduction}\label{par:intro}

Benefiting from the exponential growth of pre-training corpora and model parameters, large language models (LLMs) have achieved tremendous success in many complex application scenarios across multiple domains, including but not limited to code copilot~\cite{barke2023grounded}, clinical diagnosis~\cite{rao2023evaluating}, and market prediction~\cite{wu2023bloomberggpt}.

However, as the scale of datasets increases, data transparency is gradually declining. It has an increasing tendency for pre-training data to be treated as private and confidential rather than publicly disclosed, even for some open-source LLMs~\cite{touvron2023llama, team2024gemma}. This lack of transparency can lead to ethical concerns and pose challenges in model evaluation, if copyrighted, private or evaluation data is exposed to the target model during the pretraining phase. For instance, prior studies have found that ChatGPT has memorized a significant amount of copyrighted material~\cite{chang2023speak}, which is likely to have been used in training the GPT model, several studies show that considerable private data can be extracted from LLMs through a black-box access~\cite{carlini2021extracting, carlini2022quantifying}. Furthermore, ~\citeauthor{oren2023proving}~\citeyearpar{oren2023proving} demonstrate that the benchmark data could be included during the pre-training, leading to an overestimation of the model performance. Therefore, in this paper, we aim to investigate a problem of significant importance: detecting pre-training data of the LLM.

Pre-training data detection refers to determining whether a given pending text is included in the pre-training corpus of the target LLM, which can be considered as an instance of membership inference attack (MIA)~\cite{shokri2017membership}. Previous research on MIAs against LLMs has primarily focused on small-scale language models or fine-tuned LLMs. This is due to the unique characteristics of LLM pre-training, such as larger-scale corpora, fewer training epochs, and less knowledge of the training data distribution~\cite{shi2023detecting, duan2024membership}. For example, several studies employed reference-free attacks that identify training samples based on statistical scores evaluated on the fine-tuned model, such as perplexity (PPL)~\cite{yeom2018privacy}. Furthermore, other studies achieve higher precision through reference-based attacks, which calibrate the score with a specific ``referenced score''. Existing studies have investigated that comparing the sample PPL to zlib compression entropy~\cite{carlini2021extracting}, the lowercased sample PPL~\cite{carlini2021extracting}, and the neighboring samples PPL~\cite{mattern2023membership}. Some studies further consider comparing with the sample PPL under a reference model, such as the pre-trained model before fine-tuning~\cite{mireshghallah2022quantifying}, the smaller model has same architecture~\cite{carlini2021extracting}. ~\citeauthor{fu2023practical}~\citeyearpar{fu2023practical} fine-tune a reference model based on the output of the target model, and achieve an inspiring detection performance over fine-tuned LLMs. Recently, a benchmark dataset~\cite{shi2023detecting} and two reference-free methods~\cite{shi2023detecting, zhang2024mink} are proposed to dedicate on detecting pre-training data, which focus on token-level rather than the sentence-level likelihood.

However, despite the previous study has achieved considerable achievements in detecting pre-training data, there still exists the following limitations in the research of this problem: First, the widely adopted benchmark dataset, WIKIMIA, can only evaluate LLMs released or pre-trained before January 2023~\cite{shi2023detecting}. Over the past year, massive new state-of-the-art LLMs like LLaMA-2~\cite{touvron2023llama}, Gemma~\cite{team2024gemma} have emerged, making WIKIMIA somewhat outdated for assessing the vulnerabilities of these models to MIA. Second, with the development of AI alignment technologies, there is an increasing tendency that aligns pre-trained LLMs (unaligned) with human values and intentions~\cite{ouyang2022training}. 
However, conducting MIA on aligned LLMs is more challenging, and this remains an open problem. 
Since safety alignment will restrict the harmful behaviors of LLMs during inference~\cite{ji2024beavertails}.
Additionally, fine-tuning LLMs to align with human intentions may lead to catastrophic forgetting~\cite{luo2023investigating, luo2024empirical}, which will reduce the model memorization on pre-training data and making it more difficult to identify them~\cite{shi2022just}.
Finally, the performance of existing methods for detecting pre-training data is unsatisfactory or relies on the selection of algorithm parameters, such as $k$ for Min-K\%++~\cite{zhang2024mink}, which differ among various models, posing challenges in determining the optimal parameters in practical scenarios. 

In this paper, we first construct a more up-to-date benchmark dataset for evaluating LLMs released recently, where we fetch articles from Wikipedia event pages, then set March 1, 2024, as the cutoff date. The articles before this date will be considered as member data that had been utilized for pre-training, and the others will compose the non-member set. In addition, unlike existing methods that attempt to design various external sophisticated score functions, we propose a novel paradigm, MIA-Tuner, for pre-training data detection: internally instructing LLMs themselves to identify text that belongs to their own pre-training dataset. Specifically, we utilize the instruction-tuning~\cite{wei2021finetuned, ouyang2022training} to induce the aligned LLM to directly answer whether a pending text provided by the user belongs to their pre-training dataset. We adopt the supervised fine-tuning (SFT) to adapt the unaligned LLM to amplify the PPL discrepancy between member and non-member samples. Based on the intuition of MIA-Tuner, we also design two novel pipelines to defend LLMs against existing detection methods and the adversary version of MIA-Tuner.

Overall, our contributions are summarized as follows:

\begin{itemize}
    \item We construct a more up-to-date dataset, WIKIMIA-24, for evaluating pre-training data detection methods, which sets March 2024 as the cutoff date, allowing us to evaluate all LLMs released before that time.
    \item We propose MIA-Tuner, a novel pre-training data detection method that can persuade LLMs themselves to serve as effective and efficient pre-training text detectors. Two instances of MIA-Tuner can be applied to both aligned and unaligned LLMs. Additionally, we design two safeguards based on the intuition of MIA-Tuner to defend LLMs against both existing and proposed methods.
    \item We conducted extensive experiments to validate the effectiveness and the practicability of the MIA-Tuner. The results demonstrate that MIA-Tuner achieves significantly higher detection performance and stability across multiple aligned and unaligned LLMs compared with existing MIAs (about $53.4\%$ and $26.5\%$ improvement in AUC on aligned and unaligned LLMs, respectively).
\end{itemize}




\section{Related Works}

\para{Membership Inference Attacks}
Pre-training data detection task can be considered an instance of membership inference attack (MIA), which aims to infer whether a given sample belongs to the training set of the target model~\cite{shokri2017membership}. MIA has been well investigated across various machine learning tasks, like classification~\cite{choquette-choo2021labelonly}, recommendation~\cite{wang2022debiasing}, and generation~\cite{duan2023are, fu2023probabilistic}. The recent success of LLMs has spurred research into MIA against these models, which is of great value for quantifying privacy risks~\cite{mireshghallah2022quantifying}, detecting copyright-protected content~\cite{shi2023detecting}, and evaluating model memorization~\cite{mireshghallah2022empirical}. The prior investigation predominantly focused on fine-tuned language models~\cite{mattern2023membership, fu2023practical}. Due to the larger scale of pre-training corpora, fewer training epochs, and the inaccessibility of the training data distribution, conducting MIA against pre-trained LLMs is more challenging~\cite{zhang2024mink}. ~\citeauthor{shi2023detecting} is the pioneer to investigate this problem, who propose Min-K\% to utilize the average over the $k$ minimum token probabilities for detection. ~\citeauthor{zhang2024mink} propose Min-K\%++, an enhanced version of Min-K\%, motivated by the insight that training data tends to be around the local maximum~\cite{fu2023practical}. However, our experiments show that existing methods fail to achieve sufficiently high detection accuracy and exhibit noticeable performance degradation on aligned LLMs. Some state-of-the-art methods (e.g., Min-K\%++) are limited by model-specific, carefully designed parameters. In this work, we propose the MIA-Prompter, which instructs LLMs themselves to conduct pre-training detection with higher confidence.

\para{Fine-tuning and Alignment}
Fine-tuning has become a mainstream approach for adapting pre-trained LLMs to the downstream applications, which customizes the pre-trained model over a smaller domain-specific dataset with lower computational overhead~\cite{han2024parameter}. To further improve the effectiveness and efficiency of adaption, massive parameter-efficient fine-tuning (PEFT) approaches have been proposed, such as LoRA~\cite{hulora}, Prompt-Tuning~\cite{lester2021power}, and ${(\text{IA})^3}$~\cite{liu2022few}. Most fine-tuning techniques are deployed in a supervised manner, which refers to supervised fine-tuning (SFT).
Model alignment aims to fine-tune LLM aligning with human values and intentions through reinforcement learning based techniques like instruction tuning~\cite{weifinetuned} and supervised learning based techniques like reinforcement learning from human feedback, RLHF~\cite{ouyang2022training}. The instruction tuning can be considered as conducting SFT over an instruction dataset rather than a plain corpus~\cite{weifinetuned}.
It is worth mentioning that some concurrent studies have shown that fine-tuning aligned or unaligned LLMs may lead to privacy compromises. For instance, fine-tuning aligned LLMs can compromise safety alignment~\cite{qi2023finetuning}, while fine-tuning unaligned LLMs can further exacerbate verbatim memorization~\cite{ozdayi2023controlling, zhang2023ethicist}.
In this work, we utilize instruction tuning and supervised fine-tuning to induce aligned and unaligned LLMs themselves to detect whether a given data sample belongs to their pre-training set.

\section{Preliminary}
\subsection{Large Language Models (LLMs)}
Since currently state-of-the-art LLMs are typically deployed in an auto-regressive manner~\cite{touvron2023llama, team2024gemma}, all LLMs mentioned in this paper will refer to autoregressive LLMs unless otherwise specified. LLMs aim to predict the conditional generation probability $p_\theta \left( t_i \mid \boldsymbol{x}_{< i}\right)$ given the previous tokens $\boldsymbol{x}_{< i} = \left[t_1, t_2, \cdots, t_{i-1} \right]$. Thus, LLM can generate coherent tokens by sampling one token at a time and producing a complete text following an auto-regressive manner.
During the pre-training phase, LLMs are trained to maximize the generation probability of training texts, which can be decomposed into the product of conditional probability:
\begin{equation}\label{equ:loss}
\mathcal{L}_{\theta}(\boldsymbol{x})=- \sum_{i=1}^{\left|\boldsymbol{x}^{}\right|} \log p_\theta \left( t_i \mid \boldsymbol{x}_{< i}^{}\right).
\end{equation}
The pre-trained LLMs without alignment can only generate output patterns observed in the training data without explicit consideration for aligning user intentions~\cite{shen2023large}. In practice, methods like instruction-tuning~\cite{wei2021finetuned} and reinforcement learning from human feedback (RLHF)~\cite{ouyang2022training} are widely adopted for tuning LLMs to respond to user instruction naturally, the tuned LLMs are referred to aligned LLMs.

    

\subsection{Problem Statement and Threat Model}
The pre-training data detection task can be considered as an instance of MIA~\cite{shi2023detecting}. Thus, we provide the statement of this task by illustrating the threat model of MIA-Tuner. Given a text corpus $D$ that is split into two subsets $D_{mem}$ and $D_{non}$ and an LLM $f_\theta$ parameterized by $\theta$. The member set $D_{mem}$ is used for pre-training $f_\theta$ and $D_{non}$ is referred to as the non-member set. We consider an adversary $\mathcal{A}$ who leverages a confidence scoring function $s$ to infer whether a given text sample $\boldsymbol{x} \in D$ was seen by the target LLM $\theta$ during the pre-training phase:
\begin{equation}
    \mathcal{A}\left( \boldsymbol{x}^{}, \theta \right) = 
  \mathbb{1}\left[
  s\left(\boldsymbol{x}^{}, \theta\right) \geq \tau
  \right],
\end{equation}
where $\mathcal{A}\left( \boldsymbol{x}^{}, \theta \right)=1$ indicates that $\boldsymbol{x}^{} \in D_{mem}$, and $\tau$ denotes the tunable decision threshold. 
Like the common assumption~\cite{zhang2024mink, shi2023detecting}, we also assume the adversary can access LLM output statistics (e.g., loss value and token logits). We further extend two moderate assumptions that are feasible for all open-sourced LLMs and some commercial LLMs (e.g., ChatGPT~\cite{achiam2023gpt}): 1) the adversary is approved to directly fine-tune LLM or invoke the fine-tuning API of LLM. 2) the adversary can draw a small-scale set of member and non-member samples from common pre-training corpora (e.g., Wikipedia) based on the release date of the target LLM.

\section{Methodology}

In this section, we briefly demonstrate the motivation and intuition of MIA-Tuner before diving into the technical details. Then we propose the notion of MIA-Tuner, as well as the two pipelines to deploy it for aligned and unaligned LLMs.

\subsection{Motivation \& Intuition}

\begin{figure*}[t!]
    \centering
    {\includegraphics[width=1\textwidth]{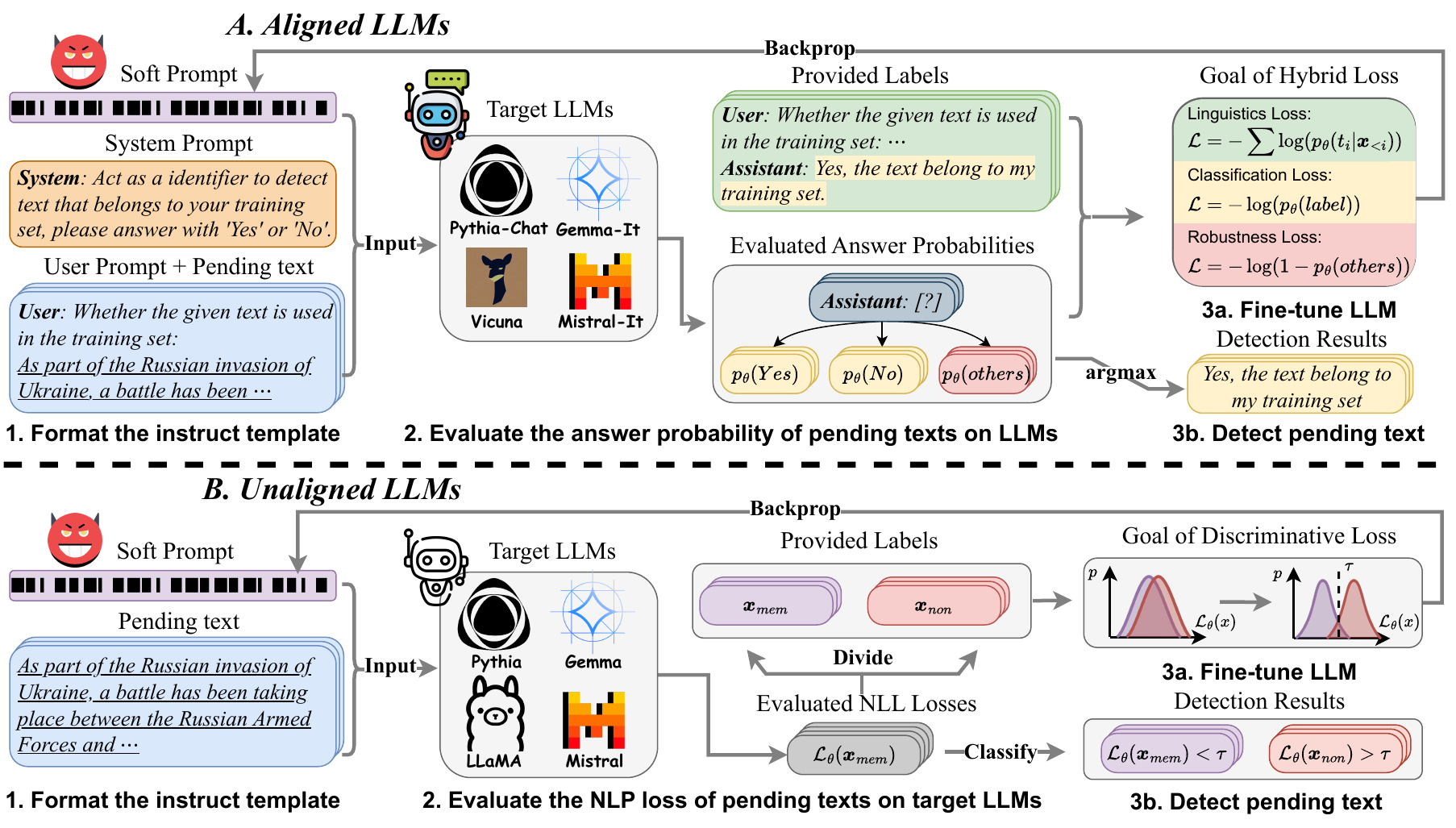}}
    \caption{The overall framework of MIA-Tuner and the two pipelines designed for aligned and unaligned LLMs, resprectively.}
    \label{fig:framework}
    \vspace{-10pt}
\end{figure*}

Existing pre-training data detection methods that mainly focus on curating or calibrating sophisticated statistical metrics have failed to provide precise, confident, and robust detection results. This renders existing methods ineffective in scenarios such as exposing LLM privacy or identifying copyright violations. To address the aforementioned issues, we raise a promising intuition before diving into in-depth technical details: \textit{Can LLMs be prompted or instructed to detect the pre-training texts by themselves?} That is, a more reliable detection method is expected by prepending a series of instructive tokens, $\boldsymbol{p} = \left[p_1, \cdots, p_{n} \right]$:
\begin{equation}
    \mathcal{A}\left( \boldsymbol{x}^{}, \theta \right) = 
  \mathbb{1}\left[
  s\left([\boldsymbol{p}; \boldsymbol{x}^{}], \theta\right) \geq \tau
  \right].
\end{equation}
We attempt to answer this question based on the success of the prompt-tuning~\cite{lester2021power} and the instruct-tuning~\cite{wei2021finetuned} in the field of LLM. Following the paradigm of prompt-tuning, we remove the restriction that the prompt be parameterized by the embedding layer of LLM; instead the prompt has its own dedicated parameters, $\phi$, that can be fine-tuned.

\subsection{Tuning LLMs to Conduct Detection} 
The overview of MIA-Tuner is illustrated in Figure~\ref{fig:framework}. In this framework, we inject and fine-tune an adversarial soft prompt to stimulate LLMs to recall memorization of the training text. Subsequently, considering the different intentions of aligned and unaligned LLMs, we design two distinct pipelines for fine-tuning aligned and unaligned LLMs, respectively. For aligned LLMs, which are already aligned with human feedback, we fully explore this characteristic to fine-tune LLMs to become pre-training text detection assistants. We use instruction fine-tuning to align LLMs with our intention of directly answering \textit{``Yes''} or \textit{``No''} the given pending text belongs to the pre-training set. Thus, the confidence scoring function can be formulated as the ratio of the probability that aligned LLM answers \textit{``Yes''} or \textit{``No''}:
\begin{equation}
    \mathcal{A}\left( \boldsymbol{x}^{}, \theta \right) = 
  \mathbb{1}\left[
 \frac{p_\theta(\textit{``Yes''} \mid [\boldsymbol{p}_\phi; \boldsymbol{p}_s; \boldsymbol{p}_u; \boldsymbol{x}])}{p_\theta(\textit{``No''} \mid [\boldsymbol{p}_\phi; \boldsymbol{p}_s; \boldsymbol{p}_u; \boldsymbol{x}])} \geq \tau
  \right],
\end{equation}
where $\boldsymbol{p}_\phi$, $\boldsymbol{p}_s$, $\boldsymbol{p}_u$ respectively denote the soft, system and user prompts, $\tau$ is set to $1$. The prompt template of each aligned LLM can be found in Appendix~\ref{par: template}.
Unlike aligned LLM, unaligned LLM cannot directly answer the pre-training text detection question. Therefore, following existing research, we use the loss as a metric to discriminate member texts and fine-tune LLM to amplify the obscured differences in this distribution, providing a clearer boundary. Thus, the scoring function is defined as the loss value of the pending text on the target LLM:
\begin{equation}
    \mathcal{A}\left( \boldsymbol{x}^{}, \theta \right) = 
  \mathbb{1}\left[
 \mathcal{L}_{\theta}([\boldsymbol{p}_\phi; \boldsymbol{x}]) \leq \tau
  \right].
\end{equation}

Finally, we elaborately design two optimization goals for fine-tuning the adversarial soft prompt injected into aligned and unaligned LLMs.

\subsection{Hybrid Loss for Aligned LLMs} 
We essentially follow the existing instruction tuning pipeline~\cite{wei2021finetuned} to train the malicious soft prompt, but made corresponding improvements tailored to our intention in the optimization goal. 
Specifically, we designed a new hybrid loss from three dimensions to ensure that aligned large language models can assist users in identifying pre-training set texts through dialogue: 1) \textbf{Linguistics}: LLM should resist basic linguistic capability to answer user questions.
2) \textbf{Classification}: LLM should be proficient in distinguishing between member and non-member texts. 3) \textbf{Robustness}: LLM should ensure the validity of output answers. We curate a hybrid loss that is composed of three parts to address the aforementioned requirements:
\begin{equation}
\mathcal{L}_{\textit{hb}}(\boldsymbol{x}^{}) = \alpha \mathcal{L}_{\textit{lg}}(\boldsymbol{x}^{}) + \beta \mathcal{L}_{\textit{cl}}(\boldsymbol{x}^{}) + \gamma \mathcal{L}_{\textit{rb}}(\boldsymbol{x}^{}),
\end{equation}
where the components $\mathcal{L}_{\textit{lg}}$, $\mathcal{L}_{\textit{cl}}$, and $\mathcal{L}_{\textit{rb}}$ correspond to the loss of the linguistics, classification, and robustness parts, respectively. $\alpha$, $\beta$, and $\gamma$ are weights of each part.

Similar to the existing instruction tuning~\cite{wei2021finetuned, wang2023selfinstruct}, we employ the commonly used negative log-likelihood (NLL) loss as the linguistics part of the hybrid loss. Most instruction-tuning solutions either mask the prompt part loss or endorse the entire sequence loss~\cite{huerta-enochian2024instruction}. To better balance the attention of the LLM between the prompt and completion parts, we re-weighted the different parts of the NLL loss:
\begin{equation}
\mathcal{L}_{\textit{lg}}(\boldsymbol{x})=- \sum_{i=1}^{\left|\boldsymbol{x}^{}\right|} w_i \log p_\theta \left( t_i \mid [\boldsymbol{p}_\phi; \boldsymbol{p}_s; \boldsymbol{p}_u; \boldsymbol{x}_{< i}^{}]\right),
\end{equation}
where $w_i$ is the loss weight of $t_i$. Following the default setting of ChatGPT~\cite{dodgson2023establishing}, $w_i$ set to $0.01$ and $1$ for prompt and completion parts, respectively.

We further adopt the cross-entropy loss as the classification part of the hybrid loss. Particularly, we first renormalize the probability that the victim aligned LLM answers \textit{``Yes''} or \textit{``No''}, then measure the negative log-likelihood of the victim LLM performs a correct answer: 
\begin{equation}
\mathcal{L}_{\textit{cl}}(\boldsymbol{x})=- \log p_\theta(label \mid [\boldsymbol{p}_\phi; \boldsymbol{p}_s; \boldsymbol{p}_u; \boldsymbol{x}]),
\end{equation}
where $label=\textit{``Yes''}$ corresponds to the pending text belongs to the pre-training dataset; otherwise, $label=\textit{``No''}$.

Furthermore, we assign a penalty value to illegal tokens other than \textit{``Yes''} or \textit{``No''} as part of the robustness of the hybrid loss:

\begin{equation}
\mathcal{L}_{\textit{rb}}(\boldsymbol{x})=- \log \left(1-p_\theta(others \mid [\boldsymbol{p}_\phi; \boldsymbol{p}_s; \boldsymbol{p}_u; \boldsymbol{x}]) \right),
\end{equation}
where $others$ refers to all illegal answer tokens.

\subsection{Contrastive Loss for Unaligned LLMs}\label{par:perturbation}
We adopt the existing fine-tuning pipeline~\cite{lester2021power} to train the malicious soft prompt, and the optimization goal is designed to amplify the discrepancy between member and non-member data with regard to the loss value. Inspired by the intuition of contrastive learning~\cite{chen2020simple}, we refer the form of NT-Xent Loss~\cite{sohn2016improved} to maximize agreement among different samples from the same class (member or non-member). Specifically, we randomly sample a batch of $2N$ samples, which includes $N$ member and $N$ non-member samples for each training batch. Given a member sample, we treat the other $N-1$ member samples as positive samples and the $N$ non-member samples as negative samples. Let $d(\boldsymbol{x}_m, \boldsymbol{x}_n) = exp(-(\mathcal{L}(\boldsymbol{x}_m) - \mathcal{L}(\boldsymbol{x}_n)))$ denotes the MIA score distance between samples $\boldsymbol{x}_m$ and $\boldsymbol{x}_n$. Thus, the loss function for a specific sample is formulated as:
\begin{equation}\label{equ:ctr_loss}
    \mathcal{L}_{\textit{ctr}}(\boldsymbol{x}_m) = - \operatorname{log}\frac{ \sum_{\boldsymbol{x}_k \in \mathcal{P}_m} exp\left(d(\boldsymbol{x}_m, \boldsymbol{x}_k)/\tau\right)}{\sum_{n = 1}^{2N} \mathbb{1}_{[n \neq i]}exp(d(\boldsymbol{x}_m, \boldsymbol{x}_n)/\tau)},
\end{equation}
where $\mathcal{P}_m$ denotes the $N-1$ positive samples of a sample $\boldsymbol{x}_m$, $\mathbb{1}_{[k \neq i]}$ is an indicator function equaling to 1 iff $[k \neq i]$, and $\tau$ represents the temperature. The overall loss is calculated over all positive pairs, both member and non-member samples, in a batch.

\subsection{Tuning LLMs to Defend Detection}\label{par:defender}
Except from conducting MIA, defending against MIA is also a topic of interest in the current community. From the opposite perspective, MIA-Tuner should also be capable of easily inducing an LLM to defend against external pre-training data detection. Therefore, we attempted to explore the possibility of reversing the optimization goals, initially designed for attacks, to meet defense requirements. Specifically, for the existing metric-based methods and the version of MIA-Tuner on unaligned LLMs, we considered narrowing the difference in loss value distribution between member and non-member samples. We modified the contrastive loss in Eq.~\ref{equ:ctr_loss} to make the distances of positive and negative pairs as similar as possible:
\begin{equation}\label{equ:def_unaligned}
    \mathcal{L}_{\textit{def}}(\boldsymbol{x}) = \left| \mathcal{L}_{\textit{ctr}}(\boldsymbol{x}) + \operatorname{log}\frac{N-1}{2N-1} \right|,
\end{equation}
where $\mathcal{L}_{\textit{ctr}}(\boldsymbol{x})$ will equal to $-\operatorname{log}\frac{N-1}{2N-1}$ when the distances of positive and negative pairs are equal.

For the version of MIA-Tuner on aligned LLMs, we only considered reversing the robust loss to guide the LLM to refuse to provide valid answers for the pre-training data detection task:

\begin{equation}\label{equ:def_aligned}
\mathcal{L}_{\textit{def}}(\boldsymbol{x})=- \log \left(p_\theta(other \mid [\boldsymbol{p}_\phi; \boldsymbol{p}_s; \boldsymbol{p}_u; \boldsymbol{x}]) \right),
\end{equation}
where $p_\theta(others \mid [\boldsymbol{p}_\phi; \boldsymbol{p}_s; \boldsymbol{p}_u; \boldsymbol{x}])$ will equal to 1 when the LLM assigns zero probability to valid answer tokens. We chose not to modify the linguistic loss because we want the defense method to have no significant impact on the LLM's language capabilities. Similarly, directly reversing the optimization of the classification loss would lead LLMs to provide valid but precisely opposite responses (\textit{``No''} for member, \textit{``Yes''} for non-member).
\section{Experiments}

\begin{table*}
\centering

\caption{Performance of MIA-Tuner across seven pre-trained LLMs with both aligned and unaligned versions. \textbf{Bold} and \colorbox{mygray}{Shade} respectively denote the best and the second-best results for each target LLM. The proposed MIA-Tuner strikes remarkable performance margins over all baselines.
N/A demonstrates that there not exists a smaller version of the target LLM for conducting Smaller Ref.}
\label{tab:attack performance}
\resizebox{\linewidth}{!}{%
\begin{tabular}{clccccclcccccccc} 
\hline
\multirow{2}{*}{\textbf{Method}} & \multicolumn{7}{c}{\textbf{Aligned LLMs}}                                                                                                                                                                                                                                                                         &  & \multicolumn{7}{c}{\textbf{Unaligned LLMs}}                                                                                                                                                                                                                                                                        \\ 
\cline{2-8}\cline{10-16}
                                 & Pythia                                    & Falcon                                    & Vicuna                                    & LLaMA-2                                   & Mistral                                   & Gemma                                     & \textbf{\textbf{Avg.}}                    &  & Pythia                                    & Falcon                                    & LLaMA                                     & LLaMA-2                                   & Mistral                                   & Gemma                                     & \textbf{Avg.}                              \\ 
\hline\hline
PPL                              & 0.693                                     & 0.617                                     & 0.654                                     & 0.614                                     & 0.571                                     & 0.520                                     & 0.612                                     &  & 0.714                                     & 0.641                                     & 0.681                                     & 0.619                                     & 0.604                                     & 0.589                                     & 0.641                                      \\
Min-K\%                          & {\cellcolor[rgb]{0.784,0.784,0.784}}0.738 & 0.644                                     & {\cellcolor[rgb]{0.784,0.784,0.784}}0.655 & {\cellcolor[rgb]{0.784,0.784,0.784}}0.642 & 0.586                                     & 0.531                                     & {\cellcolor[rgb]{0.784,0.784,0.784}}0.633 &  & {\cellcolor[rgb]{0.784,0.784,0.784}}0.759 & 0.685                                     & 0.704                                     & 0.650                                     & 0.634                                     & 0.617                                     & 0.675                                      \\
Min-K\%++                        & 0.656                                     & {\cellcolor[rgb]{0.784,0.784,0.784}}0.736 & 0.518                                     & 0.571                                     & 0.569                                     & 0.528                                     & 0.596                                     &  & 0.750                                     & {\cellcolor[rgb]{0.784,0.784,0.784}}0.831 & {\cellcolor[rgb]{0.784,0.784,0.784}}0.788 & {\cellcolor[rgb]{0.784,0.784,0.784}}0.771 & {\cellcolor[rgb]{0.784,0.784,0.784}}0.753 & {\cellcolor[rgb]{0.784,0.784,0.784}}0.756 & {\cellcolor[rgb]{0.784,0.784,0.784}}0.775  \\ 
\hline
Zlib                             & 0.716                                     & 0.643                                     & 0.678                                     & 0.631                                     & {\cellcolor[rgb]{0.784,0.784,0.784}}0.590 & 0.535                                     & 0.632                                     &  & 0.737                                     & 0.662                                     & 0.701                                     & 0.636                                     & 0.620                                     & 0.606                                     & 0.660                                      \\
Lowercase                        & 0.689                                     & 0.626                                     & 0.627                                     & 0.606                                     & 0.531                                     & {\cellcolor[rgb]{0.784,0.784,0.784}}0.553 & 0.605                                     &  & 0.694                                     & 0.636                                     & 0.651                                     & 0.625                                     & 0.594                                     & 0.615                                     & 0.636                                      \\
Neighbor                         & 0.664                                     & 0.591                                     & 0.643                                     & 0.611                                     & 0.572                                     & 0.529                                     & 0.602                                     &  & 0.671                                     & 0.624                                     & 0.659                                     & 0.617                                     & 0.602                                     & 0.579                                     & 0.625                                      \\
Smaller Ref                      & 0.629                                     & N/A                                       & N/A                                       & N/A                                       & N/A                                       & 0.484                                     & 0.557                                     &  & 0.641                                     & N/A                                       & N/A                                       & N/A                                       & N/A                                       & 0.661                                     & 0.651                                      \\ 
\hline
MIA-Tuner                     & \textbf{0.958}                            & \textbf{0.914}                            & \textbf{0.996}                            & \textbf{0.982}                            & \textbf{0.983}                            & \textbf{0.998}                            & \textbf{0.971}                            &  & \textbf{0.987}                            & \textbf{0.974}                            & \textbf{0.997}                            & \textbf{0.965}                            & \textbf{0.963}                            & \textbf{0.993}                            & \textbf{0.980}                             \\
\hline
\end{tabular}
}
\end{table*}

We conduct extensive experiments to evaluate the proposed MIA-Tuner and the seven representative baselines across six state-of-the-art aligned LLMs and their unaligned version. 

\subsection{Experimental Setup}
\subsubsection{Benchmark Datasets Construction} 
We employ the widely adopted pre-training data detection benchmark, WIKIMIA~\cite{shi2023detecting}, which is composed of articles from Wikipedia event pages. WIKIMIA assumes the target models were pre-trained before 2023, and set January 1, 2023 as the cutoff date for dividing member and non-member data. However, with the emergence of numerous state-of-the-art LLMs (e.g., Gemma~\cite{team2024gemma}, Mistral~\cite{jiang2023mistral}, and LLaMA-2~\cite{touvron2023llama}) over the past year, WIKIMIA is now somewhat outdated in evaluating these models. Thus, we follow a similar pipeline of WIKIMIA to collect a more up-to-date benchmark by moving forward the cutoff date to March 1, 2024. Specifically, we use the official API of Wikipedia to retrieve the articles that belong to the event category, then filter the events that happened after March 2024 as the member data. For member data, we follow the same setting of WIKIMIA only retrieving the articles created before 2017, since most pre-trained models were released after 2017.


\subsubsection{Target Models and Baselines}
We evaluate the performance of MIA-Tuner and all baselines against several state-of-the-art LLMs with both aligned and unaligned LLMs: Pythia-6.9B~\cite{biderman2023pythia}, Falcon-7B~\cite{falcon40b}, LLaMA-7B~\cite{touvron2023llamaa}, LLaMA-2-7B~\cite{touvron2023llama}, Mistral-7B~\cite{jiang2023mistral}, Gemma-7B~\cite{team2024gemma}. We employ seven representative MIA methods designed for LLMs, which were evaluated or proposed in previous works~\cite{shi2023detecting, zhang2024mink}. Including three reference-free methods: {PPL}~\cite{yeom2018privacy}, {Min-k\%}~\cite{shi2023detecting}, {Min-K\%++}~\cite{zhang2024mink}, and four reference-based methods: {Zlib}~\cite{carlini2021extracting}, {Lowercase}~\cite{carlini2021extracting}, {Neighbor}~\cite{mattern2023membership}, {Neighbor}~\cite{mattern2023membership}. Detailed information about the target models, baselines and implementation can be found in Appendix~\ref{par:detailed_settings}


\subsection{Attack Performance}
We first summarize the AUC score evaluated on the WIKIMIA-24 benchmark for all baselines and MIA-Tuner against seven aligned LLMs and their unaligned version in Table~\ref{tab:attack performance}. Furthermore, we present the evaluation results of Pythia, Falcon, and LLaMA on the WIKIMIA benchmark in Appendix~\ref{par:wikimia_performance} for a more comprehensive evaluation. The results demonstrate that MIA-Tuner provides a large performance margin for both aligned and unaligned LLMs with the highest average AUC score of 0.976, closing to the upper bound AUC=1.
Min-K\%++ achieves the second-best detection performance for almost all unaligned LLMs, and Min-K\% strikes the second-best average performance across all aligned LLMs. However, we notice that the performance of Min-K\%++ heavily relies on the hyperparameter $k$, the optimal values of which vary significantly across different target models, especially for aligned LLMs (see Appendix~\ref{par:differentk}).
Moreover, compared with the second-best baseline in each column (e.g. Min-K\%, Min-K\%++, and Zlib), MIA-Tuner increases the AUC score from about 0.71 average to a stunning level of 0.97, featuring a 36.7\% improvement. This further indicates the effectiveness and generalizability of MIA-Tuner in instructing LLMs themselves to conduct the pre-training data detection task. 
Furthermore, an interesting phenomenon is that all baselines suffer a considerable performance drop in aligned LLMs compared with their unaligned version. We attempt to interpret this phenomenon from the perspective of catastrophic forgetting~\cite{kirkpatrick2017overcoming}: the aligned LLMs are commonly fine-tuned over on their unaligned version though instruction tuning or RLHF, which have been investigated and will lead to a catastrophic forgetting on the pre-training data~\cite{luo2023investigating, luo2024empirical}. We believe this kind of forgetting not only affects the general knowledge of the target LLM, but also obliterates the memorization trails left by the pre-training data on the target LLM. Therefore, the distribution boundaries of the sophisticated metrics designed by existing methods between member and non-member samples become more blurred, leading to a decline in the performance of all baselines. In contrast, our method reactivates the faded memorization signals by instructing the LLM, thereby achieving similar performance on the aligned LLM as on its unaligned version.

\subsection{MIA-Tuner Can Be a Few-shot Learner}

\begin{figure}[t!]
    \centering
    {\includegraphics[width=.5\textwidth]{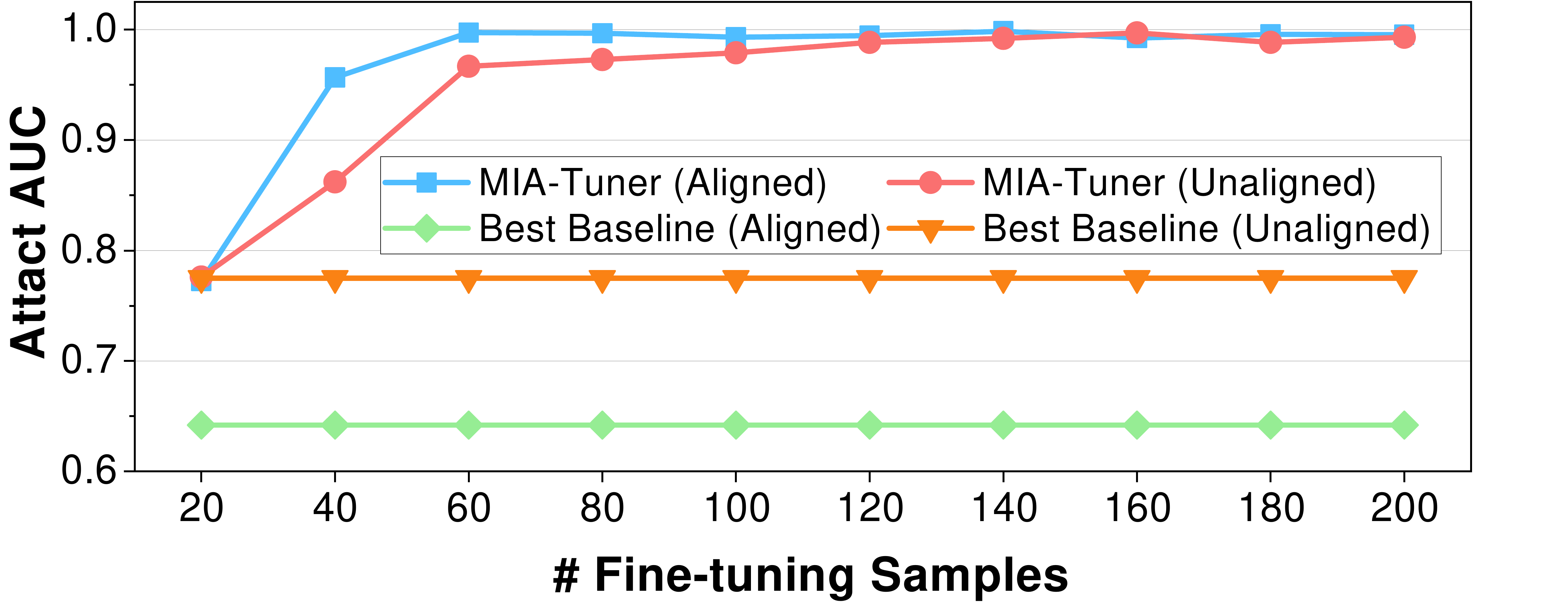}}
    \caption{The performance of MIA-Tuner on LLaMA-2 while utilizing different numbers of fine-tuning samples.}
    \label{fig:training_samples}
    \vspace{-10pt}
\end{figure}

While MIA-Tuner demonstrates a notable performance improvement in the primary evaluation compared to existing baselines, the proposed method still necessitates an additional assumption of requiring small-scale sets of member and non-member samples. One potential approach to relax this assumption could involve drawing ground truth data from a common pre-training data source. However, it is crucial to note that the scale of the data used for instructing target LLMs remains a critical factor for deploying MIA-Tuner in practical scenarios. Thus, we conducted an experiment to examine the extent to which the performance of the MIA-Tuner is dependent on the number of samples used for fine-tuning. The results presented in Figure~\ref{fig:training_samples} demonstrate that MIA-Tuner can achieve substantial detection performance on both aligned and unaligned LLMs with only 60 fine-tuning data points (30 samples for both member and non-member), with an AUC exceeding 0.95. Thus, MIA-Tuner is a few-shot learner, which can be applied with high feasibility.

\subsection{MIA-Tuner Can Be a Benign Defender}

\begin{figure}[t!]
    \centering
    {\includegraphics[width=.5\textwidth]{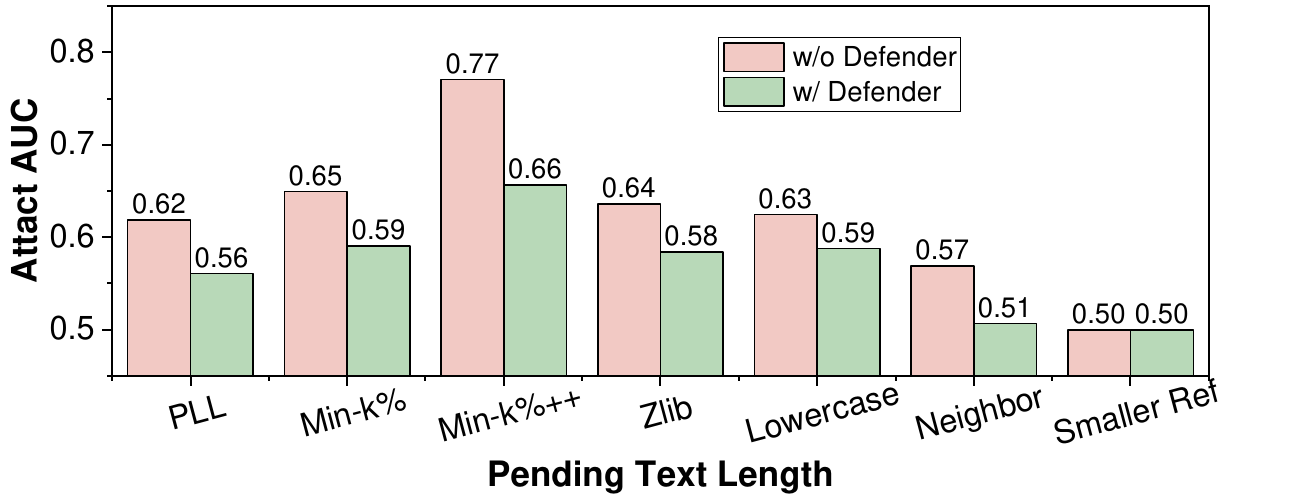}}
    \caption{The detection performance of all baselines on LLaMA-2 w/ and w/o the proposed safeguard.}\label{fig: defend_baselines}
    \vspace{-10pt}
\end{figure}

\begin{figure}[t!]
    \centering
    {\includegraphics[width=.5\textwidth]{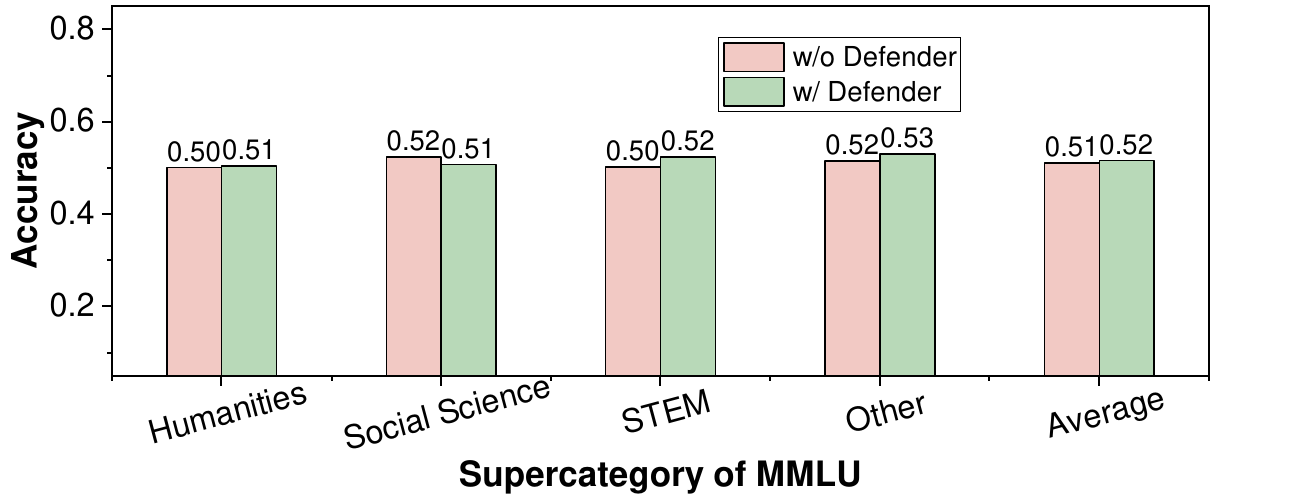}}
    \caption{The accuracy of LLaMA-2 on the MMLU benchmark w/ and w/o the proposed safeguard across four different types of tasks.}\label{fig: defend_mmlu}
    \vspace{-15pt}
\end{figure}

The existing experimental results are sufficient to demonstrate that MIA-Tuner can instruct an LLM itself to act as an attacker to perform pre-training data detection tasks. However, it remains uncertain whether MIA-Tuner can serve as an effective defender. Thus, we consider to employ the two defense methods proposed in Section~\ref{par:defender} against all existing pre-training data detection methods and the adversarial version of MIA-Tuner in two practical scenarios:

\para{1) Defend existing metric-based methods before releasing a pre-trained LLM:} Since existing methods are designed for pre-trained LLMs and have exposed considerable privacy leakage, it is essential to implement safeguards before releasing pre-trained LLMs. Thus, we utilize Eq.~\ref{equ:def_unaligned} to fine-tune a privacy-preserving model over the pre-trained model. Then, we evaluate the detection performance of all baselines on both the original and the privacy-preserving models, the results are summarized in Figure~\ref{fig: defend_baselines}. Additionally, to assess how this safeguard affects the general knowledge acquired by the LLM during the pre-training phase, we used a widely recognized benchmark, MMLU, which covers 57 categories across STEM, the humanities, the social sciences, and more~\cite{hendrycksmeasuring}. As shown in Figure~\ref{fig: defend_mmlu}, we summarized the accuracy of the original and the privacy-protected models in answering questions on 4 supercategories. The raw evaluation results on MMLU can be found in Appendix~\ref{par:mmlu}. 
As illustrated in Figure~\ref{fig: defend_baselines}, compared to the original LLM, the detection performance of all baselines on the privacy-preserving LLM has significantly decreased, with the average AUC dropping from 0.626 to 0.570. The results in Figure~\ref{fig: defend_mmlu} demonstrate that the designed safeguard can effectively prevent the pre-training data of the released LLM from being identified by existing algorithms, with almost no performance decline.

\begin{figure}[t!]
    \vspace{-2pt}
    \centering
    \includegraphics[width=.45\textwidth]{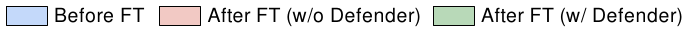}
    \vspace{1pt}
    \subfigure[Benign User]
    {\includegraphics[width=.505\linewidth]{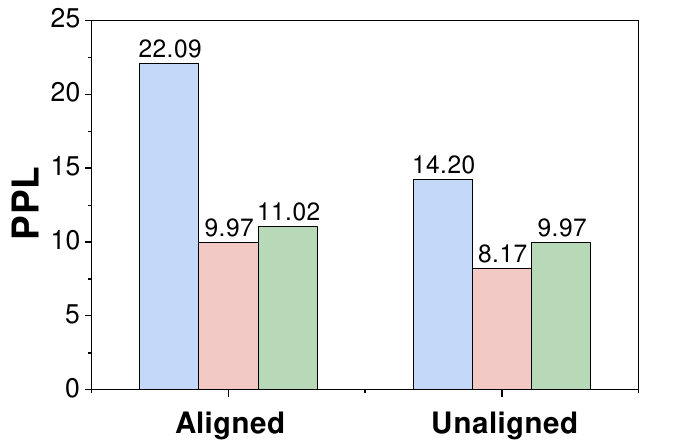}}
    \hspace{-8pt}
    \subfigure[Malicious User]
    {\includegraphics[width=.505\linewidth]{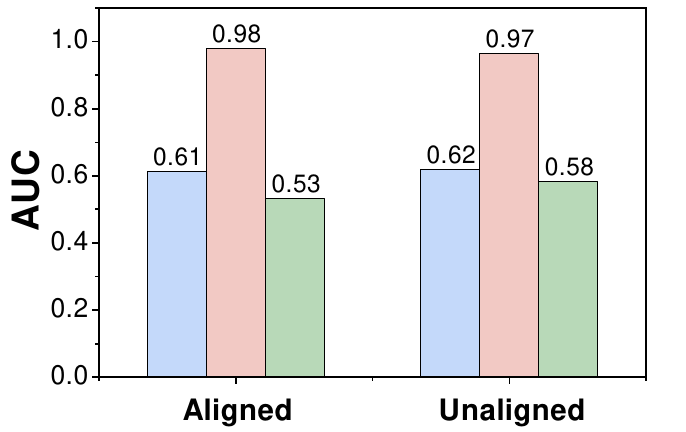}}
    \vspace{-8pt}
    \caption{The fine-tuning (FT) PPL of (a) the benign user and the detection AUC of (b) the malicious user across aligned and unaligned LLMs in three stages: Before FT, After FT (w/o Defender), After FT (w/ Defender).}
    \label{fig:defender_our}
    \vspace{-12pt}
\end{figure}

\para{2) Defend the proposed MIA-Tuner during exposing a fine-tuning API:} LLMs are often made available to the public in the form of APIs, rather than releasing pre-trained models directly. Our proposed MIA-Tuner can leverage the fine-tuning API provided by the LLM to induce it to perform pre-training data detection tasks, achieving significantly high accuracy. Therefore, we believe it is necessary to design a safeguard during the fine-tuning stage that can protect against malicious users employing MIA-Tuner, while minimizing the impact on benign users utilizing the fine-tuning API. Thus, we suggest to conduct a safeguard by fine-tuning LLMs with Eq.~\ref{equ:def_aligned} or Eq.~\ref{equ:def_unaligned} after the user fine-tuning.
We consider a malicious user and a benign user who uploads an adversary and normal fine-tuning datasets, respectively. Then we evaluate the detection AUC achieved by malicious users and the fine-tuning PPL of benign users in three stages: before fine-tuning, after fine-tuning (w/o defender), and after fine-tuning (w/ defender). As shown in Figure~\ref{fig:defender_our}, the detection performance of malicious users significantly decreased after implementing the proposed safeguard, even falling below the level before executing MIA-Tuner. In contrast, for benign fine-tuning users, the performance of the fine-tuned LLM on their customized dataset showed only a slight decline.

\subsection{MIA-Tuner Can Be a Universal Attacker}

\begin{figure}[t!]
    \vspace{-2pt}
    \centering
    \includegraphics[width=.45\textwidth]{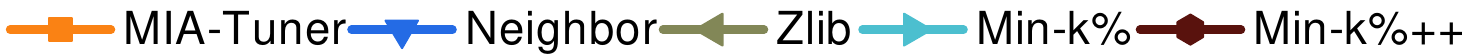}
    \vspace{1pt}
    \subfigure[Aligned LLMs]
    {\includegraphics[height=.185\textwidth]{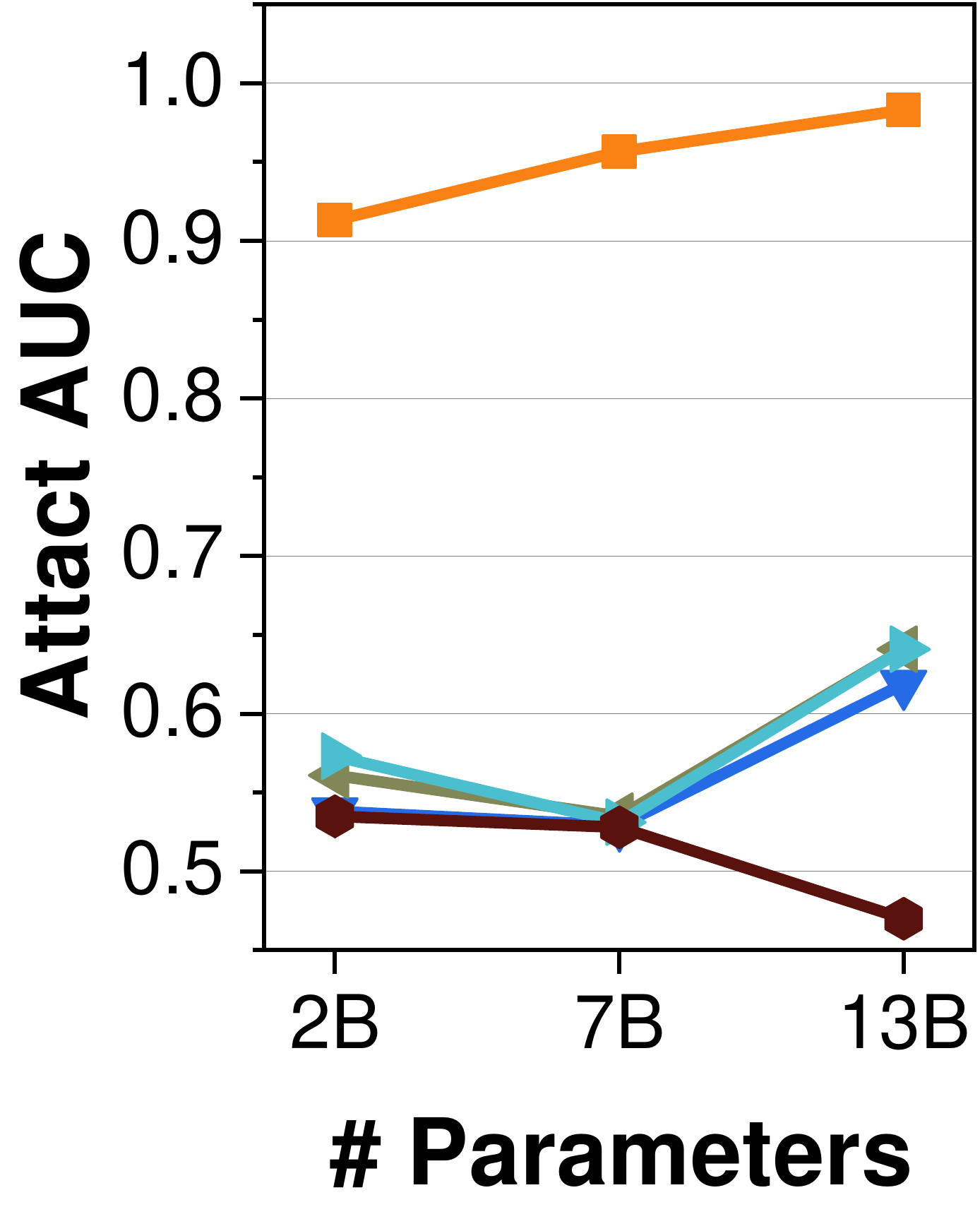}}
    \subfigure[Unaligned LLMs]
    {\includegraphics[height=.185\textwidth]{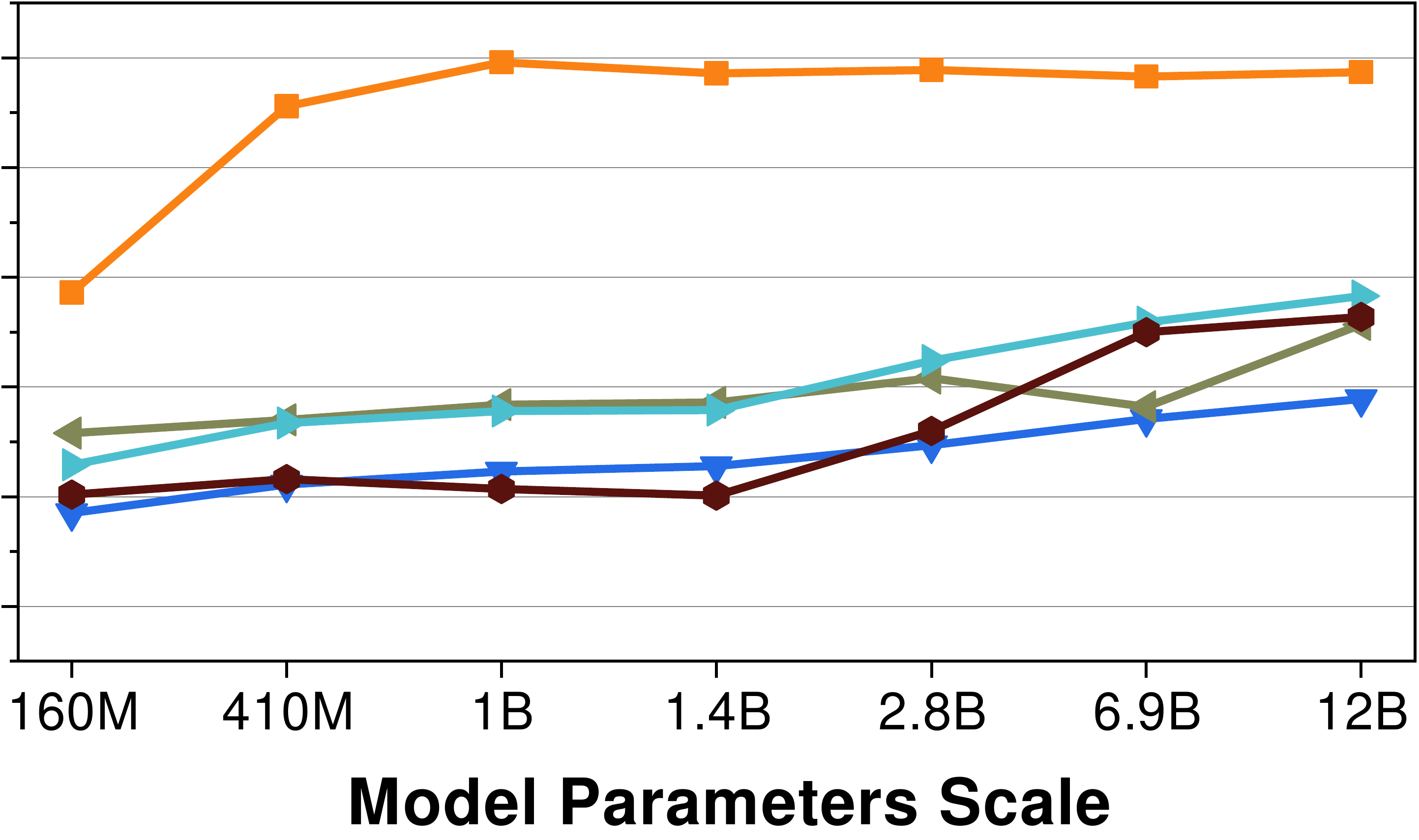}}
    \vspace{-8pt}
    \caption{The detection performance of MIA-Tuner and four representative baselines on aligned and unaligned LLMs with different parameter scales.}
    \label{fig:model_size}
    \vspace{-12pt}
\end{figure}

To further evaluate the generalizability of MIA-Tuner, we conduct experiments to investigate the two factors influencing detection difficulty: 1) the parameter scale of the target model, and 2) the length of the pending text.

\para{Parameter Scale} We evaluate the performance of MIA-Tuner and four representative baselines on detecting pre-training 128-length texts from LLMs with different scales. We adopt Gemma-(2B, 7B), and LLaMA-2-13B as aligned LLMs, Pythia-(160M, 410M, 1B, 1.4B, 2.8B, 6.9B, 12B) as unaligned LLMs. 
As shown in Figure~\ref{fig:model_size}, the AUC scores of different methods increase as the model scale increases. This may be because the larger the model scale, the more pre-training data it can memorize, and the deeper the memorization tails that detection algorithms can perceive. The reason for the performance fluctuations of Min-K\% is that the performance is overly sensitive to the parameter $k$, and the optimal $k$ value varies across different models.

\begin{figure}[t!]
    \vspace{-2pt}
    \centering
    \includegraphics[width=.45\textwidth]{images/legend_ext.pdf}
    \vspace{1pt}
    \subfigure[Aligned LLMs]
    {\includegraphics[width=.48\linewidth]{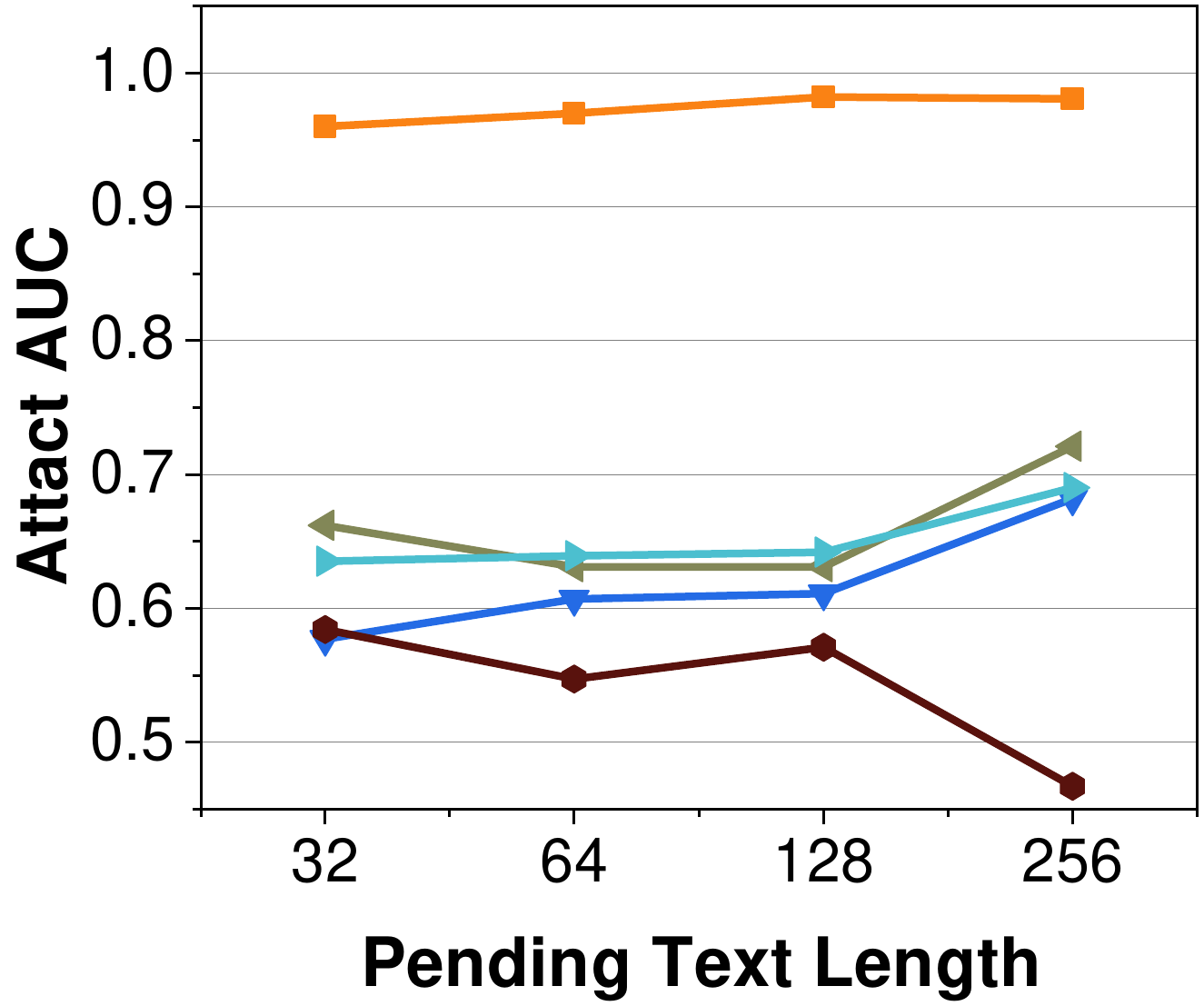}}
    \subfigure[Unaligned LLMs]
    {\includegraphics[width=.48\linewidth]{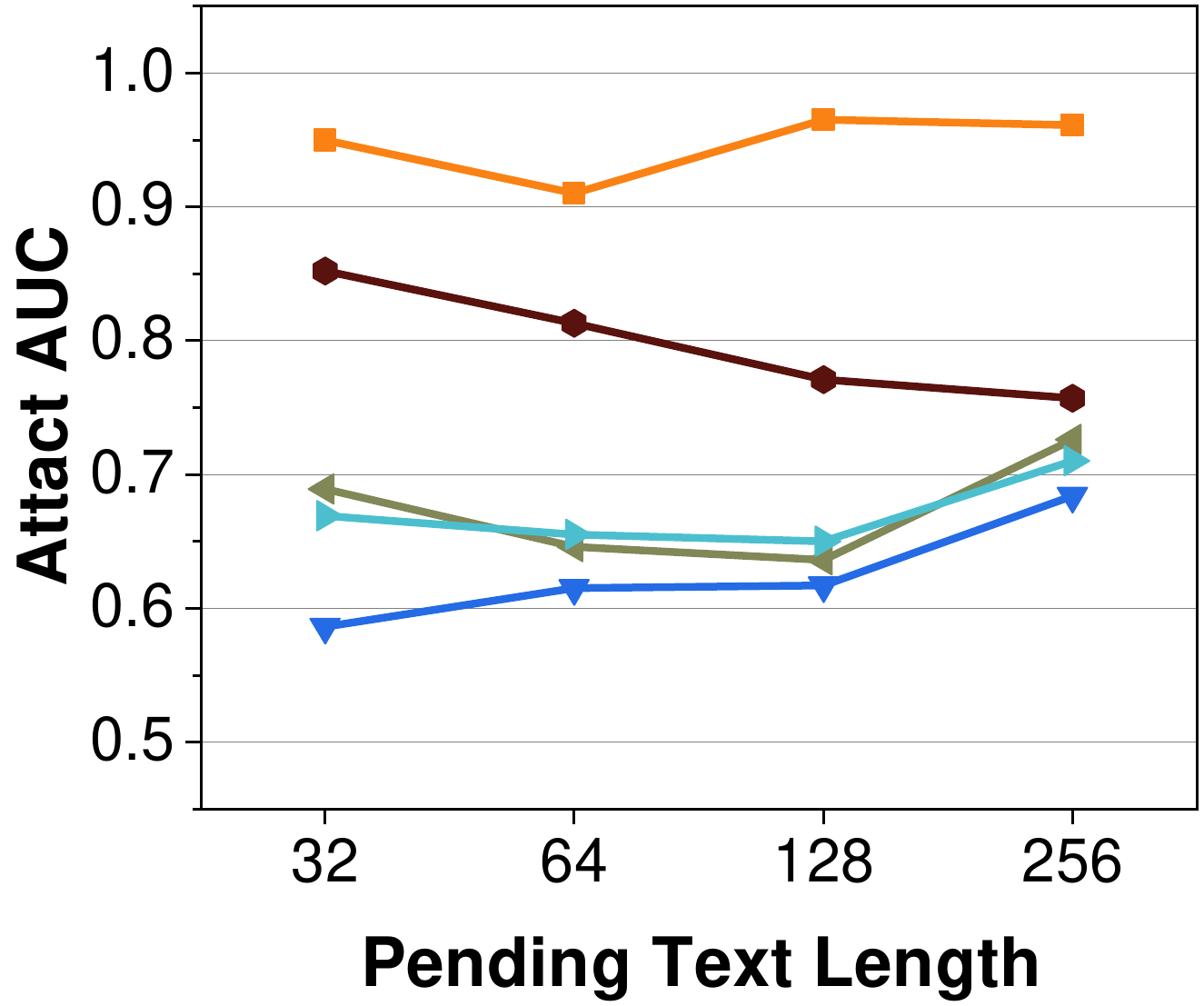}}
    \vspace{-8pt}
    \caption{
    Attack performances}
    \label{fig:text_length}
    \vspace{-12pt}
\end{figure}

\para{Text Length} We also investigate how the detection performance varies with different pending text lengths of 32, 64, 128, and 256. As shown in Figure~\ref{fig:text_length}, the detection performance gradually improves with increasing text length, suggesting that longer texts may contain more memorization tails, making them easier to identify.

\section{Conclusion}
In this article, we introduce a novel pre-training data detection dataset, WIKIMIA-24, and a new detection approach named MIA-Tuner. Our paradigm is founded on the concept that LLMs themselves can be directed to perform pre-training data detection tasks. Extensive experiments illustrate that MIA-Tuner achieves exceptionally high detection accuracy on both aligned and unaligned LLMs, surpassing existing baselines. Furthermore, MIA-Tuner operates as a few-shot attacker, necessitating only a minimal number of labeled samples to achieve satisfactory results. It also exhibits consistent detection performance across various model sizes and text lengths. Additionally, we have developed two defense strategies inspired by the MIA-Tuner framework to counter both existing methods and our proposed approach during the pre-training and fine-tuning phases, respectively. These defense strategies effectively mitigate the risk of pre-training text detection while minimally affecting the overall performance of the LLM.



\bibliography{aaai24}
\section*{Reproducibility Checklist}

This paper:

\begin{itemize}
    \item Includes a conceptual outline and/or pseudocode description of AI methods introduced (yes)
    \item Clearly delineates statements that are opinions, hypothesis, and speculation from objective facts and results (yes)
    \item Provides well marked pedagogical references for less-familiare readers to gain background necessary to replicate the paper (yes)
    \item Does this paper make theoretical contributions? (no)
\end{itemize}

If yes, please complete the list below.
\begin{itemize}
\item All assumptions and restrictions are stated clearly and formally. (yes)
\item All novel claims are stated formally (e.g., in theorem statements). (yes)
\item Proofs of all novel claims are included. (yes)
\item Proof sketches or intuitions are given for complex and/or novel results. (yes)
\item Appropriate citations to theoretical tools used are given. (yes)
\item All theoretical claims are demonstrated empirically to hold. (yes)
\item All experimental code used to eliminate or disprove claims is included. (yes)
\item Does this paper rely on one or more datasets? (yes)
\end{itemize}

If yes, please complete the list below.
\begin{itemize}
\item A motivation is given for why the experiments are conducted on the selected datasets (yes)
\item All novel datasets introduced in this paper are included in a data appendix. (yes)
\item All novel datasets introduced in this paper will be made publicly available upon publication of the paper with a license that allows free usage for research purposes. (yes)
\item All datasets drawn from the existing literature (potentially including authors’ own previously published work) are accompanied by appropriate citations. (yes)
\item All datasets drawn from the existing literature (potentially including authors’ own previously published work) are publicly available. (yes)
\item All datasets that are not publicly available are described in detail, with explanation why publicly available alternatives are not scientifically satisficing. (NA)
\item Does this paper include computational experiments? (yes)
\end{itemize}

If yes, please complete the list below.
\begin{itemize}
\item Any code required for pre-processing data is included in the appendix. (yes).
\item All source code required for conducting and analyzing the experiments is included in a code appendix. (yes)
\item All source code required for conducting and analyzing the experiments will be made publicly available upon publication of the paper with a license that allows free usage for research purposes. (yes)
\item All source code implementing new methods have comments detailing the implementation, with references to the paper where each step comes from (yes)
\item If an algorithm depends on randomness, then the method used for setting seeds is described in a way sufficient to allow replication of results. (yes)
\item This paper specifies the computing infrastructure used for running experiments (hardware and software), including GPU/CPU models; amount of memory; operating system; names and versions of relevant software libraries and frameworks. (yes)
\item This paper formally describes evaluation metrics used and explains the motivation for choosing these metrics. (yes)
\item This paper states the number of algorithm runs used to compute each reported result. (yes)
\item Analysis of experiments goes beyond single-dimensional summaries of performance (e.g., average; median) to include measures of variation, confidence, or other distributional information. (no, due to the high computational cost of the experiments, all experimental results were obtained using a fixed random seed.)
\item The significance of any improvement or decrease in performance is judged using appropriate statistical tests (e.g., Wilcoxon signed-rank). (no, the performance is significant enough)
\item This paper lists all final (hyper-)parameters used for each model/algorithm in the paper’s experiments. (yes)
\item This paper states the number and range of values tried per (hyper-) parameter during development of the paper, along with the criterion used for selecting the final parameter setting. (yes)
\end{itemize}

\clearpage
\appendix
\linespread{1}
\section{Appendix}

\subsection{Chat Templates for Aligned LLMs}\label{par: template}
\exbox{Pythia Chat Template:}
{\textbf{answer\_choice:} "Yes" (or "No")\newline
\textbf{pending\_text}: "The 2016 Boston Marathon was the 120th running of the Boston Athletic Association's mass-participation marathon."\newline
```\textit{jinja}\newline
\underline{<human>}: Please tell me whether the given example is used in the training dataset:\newline
\{\{ \textbf{pending\_text} \}\}\newline
\underline{<bot>}:\newline
\{\{ \textbf{answer\_choice} \}\}\newline
```
}

\exbox{Falcon Chat Template:}
{\textbf{answer\_choice:} "Yes" (or "No")\newline
\textbf{pending\_text}: "The 2016 Boston Marathon was the 120th running of the Boston Athletic Association's mass-participation marathon."\newline
```jinja\newline
\underline{User}: Please tell me whether the given example is used in the training dataset: \newline
\{\{ \textbf{pending\_text} \}\}\newline
\newline
\underline{Assistant}: \{\{ \textbf{answer\_choice} \}\}\newline
```
}

\exbox{Vicuna Chat Template:}
{\textbf{answer\_choice:} "Yes" (or "No")\newline
\textbf{pending\_text}: "The 2016 Boston Marathon was the 120th running of the Boston Athletic Association's mass-participation marathon."\newline
```jinja\newline
<s>
A chat between a curious user and an text identification assistant. The assistant gives certain identification answer ('Yes' or 'No') to the user provided pending text.\newline
\underline{USER}: Please tell me whether the given example is used in the training dataset: \newline
\{\{ \textbf{pending\_text} \}\}\newline
\underline{ASSISTANT}: \{\{ \textbf{answer\_choice} \}\}</s>\newline
```
}

\newpage

\exbox{LLaMA-2 Chat Template:}
{\textbf{answer\_choice:} "Yes" (or "No")\newline
\textbf{pending\_text}: "The 2016 Boston Marathon was the 120th running of the Boston Athletic Association's mass-participation marathon."\newline
```jinja\newline
<s>\underline{{[INST]}} \underline{<<SYS>>}\newline
Act as a identifier to detect text that belongs to your training set, please answer with 'Yes' or 'No' \newline
\underline{<</SYS>>}\newline
\newline
Please tell me whether the given example is used in the training dataset: \newline
\{\{ \textbf{pending\_text} \}\} \underline{[/INST]} \{\{ \textbf{answer\_choice} \}\} </s>\newline
```
}

\exbox{Mistral Chat Template:}
{\textbf{answer\_choice:} "Yes" (or "No")\newline
\textbf{pending\_text}: "The 2016 Boston Marathon was the 120th running of the Boston Athletic Association's mass-participation marathon."\newline
```jinja\newline
<s> \underline{[INST]} Act as a identifier to detect text that belongs to your training set, please answer with 'Yes' or 'No' \newline
\newline
Please tell me whether the given example is used in the training dataset: \newline
\{\{ \textbf{pending\_text} \}\} \underline{[/INST]} \{\{ \textbf{answer\_choice} \}\}</s>\newline
```
}

\exbox{Gemma Chat Template:}
{\textbf{answer\_choice:} "Yes" (or "No")\newline
\textbf{pending\_text}: "The 2016 Boston Marathon was the 120th running of the Boston Athletic Association's mass-participation marathon."\newline
```jinja\newline
<bos><start\_of\_turn>\underline{user}\newline
Please tell me whether the given example is used in the training dataset: \newline
\{\{ \textbf{pending\_text} \}\}<end\_of\_turn>\newline
<start\_of\_turn>\underline{model}\newline
\{\{ \textbf{answer\_choice} \}\}<end\_of\_turn>\newline
```
}

\begin{table*}[th]
\footnotesize
    \begin{center}
        \caption{The detailed information of target models.}\label{tab:target_llm}
\resizebox{1\linewidth}{!}{%
\begin{tabular}{lllll} 
\hline
Model Name & CardName                 & \# Para & URL                                                         & Pre-trained/Released Date  \\ 
\hline
Pythia     & Pythia-Chat-Base-7B      & 7B      & https://huggingface.co/togethercomputer/Pythia-Chat-Base-7B & March 2023                 \\
Pythia     & pythia-6.9b              & 7B      & https://huggingface.co/EleutherAI/pythia-6.9b               & Feb 2023                   \\
Falcon     & falcon-7b-instruct~      & 7B      & https://huggingface.co/tiiuae/falcon-7b-instruct            & Dec 2023                   \\
Falcon     & falcon-7b~               & 7B      & https://huggingface.co/tiiuae/falcon-7b                     & Dec 2022                   \\
Vicuna     & vicuna-7b-v1.1           & 7B      & https://huggingface.co/lmsys/vicuna-7b-v1.1                 & March 2022                 \\
LLaMA      & llama-7b                 & 7B      & https://huggingface.co/huggyllama/llama-7b                  & February 2023              \\
LLaMA-2    & Llama-2-7b-chat-hf       & 7B      & https://huggingface.co/meta-llama/Llama-2-7b-chat-hf        & July 2023                  \\
LLaMA-2    & Llama-2-7b-hf            & 7B      & https://huggingface.co/meta-llama/Llama-2-7b-hf             & July 2023                  \\
Mistral    & Mistral-7B-Instruct-v0.1 & 7B      & https://huggingface.co/mistralai/Mistral-7B-Instruct-v0.1   & Sep 2023                   \\
Mistral    & Mistral-7B-v0.1          & 7B      & https://huggingface.co/mistralai/Mistral-7B-v0.1            & Sep 2023                   \\
Gemma      & google/gemma-7b          & 7B      & https://huggingface.co/google/gemma-7b                      & Feb 2024                   \\
Gemma      & google/gemma-7b-it       & 7B      & https://huggingface.co/google/gemma-7b-it                   & Feb 2024                   \\
\hline
\end{tabular}
}
    \end{center}
\end{table*}

\begin{table*}
\centering
\caption{Performance of MIA-Tuner and all baselines across seven pre-trained LLMs with both aligned and unaligned versions. \textbf{Bold} and \colorbox{mygray}{Shade} respectively denote the best and the second-best results for each target LLM. The proposed MIA-Tuner strikes remarkable performance margins over all baselines.
N/A demonstrates that there not exists a smaller version of the target LLM for conducting Smaller Ref.}\label{tab:wikimia_performance}
\resizebox{0.8\linewidth}{!}{%
\begin{tabular}{cccccccccc} 
\hline
\multirow{2}{*}{\textbf{Method}} & \multicolumn{4}{c}{\textbf{Aligned LLMs}}                                                                                                                                     &  & \multicolumn{4}{c}{\textbf{Unaligned LLMs}}                                                                                                                                    \\ 
\cline{2-5}\cline{7-10}
                                 & Pythia                                    & Falcon                                    & Vicuna                                    & \textbf{Avg.}                             &  & Pythia                                    & Falcon                                    & LLaMA                                     & \textbf{Avg.}                              \\ 
\hline\hline
PPL                              & 0.649                                     & 0.589                                     & 0.641                                     & 0.626                                     &  & 0.651                                     & 0.611                                     & 0.666                                     & 0.643                                      \\
Min-K\%                          & {\cellcolor[rgb]{0.784,0.784,0.784}}0.689 & 0.616                                     & 0.658                                     & {\cellcolor[rgb]{0.784,0.784,0.784}}0.654 &  & {\cellcolor[rgb]{0.784,0.784,0.784}}0.695 & 0.636                                     & 0.697                                     & 0.676                                      \\
Min-K\%++                        & 0.647                                     & {\cellcolor[rgb]{0.784,0.784,0.784}}0.692 & 0.547                                     & 0.629                                     &  & 0.673                                     & {\cellcolor[rgb]{0.784,0.784,0.784}}0.783 & {\cellcolor[rgb]{0.784,0.784,0.784}}0.850 & {\cellcolor[rgb]{0.784,0.784,0.784}}0.769  \\ 
\hline
Zlib                             & 0.663                                     & 0.615                                     & {\cellcolor[rgb]{0.784,0.784,0.784}}0.661 & 0.646                                     &  & 0.675                                     & 0.628                                     & 0.683                                     & 0.662                                      \\
Lowercase                        & 0.612                                     & 0.526                                     & 0.567                                     & 0.568                                     &  & 0.605                                     & 0.560                                     & 0.591                                     & 0.583                                      \\
Neighbor                         & 0.657                                     & 0.592                                     & 0.620                                     & 0.623                                     &  & 0.676                                     & 0.627                                     & 0.659                                     & 0.654                                      \\
Smaller Ref                      & 0.628                                     & N/A                                       & N/A                                       & 0.628                                     &  & 0.633                                     & N/A                                       & N/A                                       & 0.633                                      \\ 
\hline
MIA-Prompter                     & \textbf{0.813}                            & \textbf{0.913}                            & \textbf{0.956}                            & \textbf{0.894}                            &  & \textbf{0.905}                            & \textbf{0.921}                            & \textbf{0.870}                            & \textbf{0.899}                             \\
\hline
\end{tabular}
}
\end{table*}

\begin{table*}
\centering

\caption{Performance of Min-K\% and Min-K\%++ on WikiMIA-24 with different $k\%$.}\label{tab:differentk}
\resizebox{\linewidth}{!}{%
\begin{tabular}{cccccccccccccccc} 
\hline
\multirow{2}{*}{Method} & \multicolumn{7}{c}{Aligned LLMs}                                                                                                                                                                                                                                                                                  &  & \multicolumn{7}{c}{Unaligned LLMs}                                                                                                                                                                                                                                                                                 \\ 
\cline{2-8}\cline{10-16}
                        & Pythia                                    & Falcon                                    & Vicuna                                    & LLaMA-2                                   & Mistral                                   & Gemma                                     & Avg.                                      &  & Pythia                                    & Falcon                                    & LLaMA                                     & LLaMA-2                                   & Mistral                                   & Gemma                                     & Avg.                                       \\ 
\hline\hline
Min-10\%               & 0.729                                     & 0.638                                     & 0.633                                     & 0.631                                     & 0.578                                     & 0.515                                     & 0.621                                     &  & 0.749                                     & {\cellcolor[rgb]{0.784,0.784,0.784}}0.690 & 0.685                                     & 0.648                                     & 0.631                                     & 0.611                                     & 0.669                                      \\
Min-20\%               & {\cellcolor[rgb]{0.784,0.784,0.784}}0.738 & {\cellcolor[rgb]{0.784,0.784,0.784}}0.644 & 0.655                                     & {\cellcolor[rgb]{0.784,0.784,0.784}}0.642 & {\cellcolor[rgb]{0.784,0.784,0.784}}0.586 & {\cellcolor[rgb]{0.784,0.784,0.784}}0.531 & {\cellcolor[rgb]{0.784,0.784,0.784}}0.633 &  & {\cellcolor[rgb]{0.784,0.784,0.784}}0.759 & 0.685                                     & {\cellcolor[rgb]{0.784,0.784,0.784}}0.704 & {\cellcolor[rgb]{0.784,0.784,0.784}}0.650 & {\cellcolor[rgb]{0.784,0.784,0.784}}0.634 & {\cellcolor[rgb]{0.784,0.784,0.784}}0.617 & {\cellcolor[rgb]{0.784,0.784,0.784}}0.675  \\
Min-30\%               & 0.729                                     & 0.644                                     & {\cellcolor[rgb]{0.784,0.784,0.784}}0.658 & 0.633                                     & 0.585                                     & 0.529                                     & 0.630                                     &  & 0.753                                     & 0.675                                     & 0.699                                     & 0.640                                     & 0.623                                     & 0.612                                     & 0.667                                      \\
Min-40\%               & 0.720                                     & 0.636                                     & 0.657                                     & 0.624                                     & 0.580                                     & 0.524                                     & 0.624                                     &  & 0.741                                     & 0.663                                     & 0.692                                     & 0.631                                     & 0.614                                     & 0.601                                     & 0.657                                      \\
Min-50\%               & 0.711                                     & 0.626                                     & 0.655                                     & 0.617                                     & 0.575                                     & 0.522                                     & 0.618                                     &  & 0.732                                     & 0.652                                     & 0.685                                     & 0.625                                     & 0.609                                     & 0.595                                     & 0.650                                      \\
Min-60\%               & 0.701                                     & 0.621                                     & 0.654                                     & 0.615                                     & 0.572                                     & 0.521                                     & 0.614                                     &  & 0.723                                     & 0.645                                     & 0.682                                     & 0.622                                     & 0.606                                     & 0.591                                     & 0.645                                      \\
Min-70\%               & 0.697                                     & 0.618                                     & 0.654                                     & 0.614                                     & 0.571                                     & 0.519                                     & 0.612                                     &  & 0.719                                     & 0.642                                     & 0.681                                     & 0.620                                     & 0.604                                     & 0.589                                     & 0.643                                      \\
Min-80\%               & 0.694                                     & 0.617                                     & 0.654                                     & 0.614                                     & 0.571                                     & 0.520                                     & 0.612                                     &  & 0.715                                     & 0.641                                     & 0.680                                     & 0.619                                     & 0.604                                     & 0.589                                     & 0.641                                      \\
Min-90\%               & 0.693                                     & 0.617                                     & 0.655                                     & 0.614                                     & 0.571                                     & 0.521                                     & 0.612                                     &  & 0.714                                     & 0.641                                     & 0.680                                     & 0.620                                     & 0.604                                     & 0.589                                     & 0.641                                      \\
Min-100\%               & 0.693                                     & 0.617                                     & 0.654                                     & 0.614                                     & 0.571                                     & 0.520                                     & 0.612                                     &  & 0.714                                     & 0.641                                     & 0.681                                     & 0.619                                     & 0.604                                     & 0.589                                     & 0.641                                      \\ 
\hline
Min-10\%++             & 0.656                                     & 0.736                                     & 0.518                                     & 0.571                                     & 0.569                                     & 0.528                                     & 0.596                                     &  & 0.750                                     & {\cellcolor[rgb]{0.784,0.784,0.784}}0.831 & 0.788                                     & {\cellcolor[rgb]{0.784,0.784,0.784}}0.771 & {\cellcolor[rgb]{0.784,0.784,0.784}}0.753 & {\cellcolor[rgb]{0.784,0.784,0.784}}0.756 & {\cellcolor[rgb]{0.784,0.784,0.784}}0.775  \\
Min-20\%++              & 0.682                                     & {\cellcolor[rgb]{0.784,0.784,0.784}}0.744 & 0.551                                     & 0.593                                     & 0.592                                     & {\cellcolor[rgb]{0.784,0.784,0.784}}0.529 & 0.615                                     &  & 0.773                                     & 0.817                                     & {\cellcolor[rgb]{0.784,0.784,0.784}}0.800 & 0.759                                     & 0.736                                     & 0.743                                     & 0.771                                      \\
Min-30\%++              & 0.692                                     & 0.740                                     & 0.570                                     & 0.603                                     & 0.592                                     & 0.527                                     & {\cellcolor[rgb]{0.784,0.784,0.784}}0.621 &  & 0.776                                     & 0.803                                     & 0.797                                     & 0.753                                     & 0.728                                     & 0.731                                     & 0.765                                      \\
Min-40\%++              & 0.696                                     & 0.738                                     & {\cellcolor[rgb]{0.784,0.784,0.784}}0.573 & {\cellcolor[rgb]{0.784,0.784,0.784}}0.606 & 0.592                                     & 0.524                                     & 0.621                                     &  & 0.776                                     & 0.798                                     & 0.795                                     & 0.752                                     & 0.726                                     & 0.725                                     & 0.762                                      \\
Min-50\%++              & 0.696                                     & 0.730                                     & 0.572                                     & 0.606                                     & 0.590                                     & 0.525                                     & 0.620                                     &  & {\cellcolor[rgb]{0.784,0.784,0.784}}0.777 & 0.791                                     & 0.792                                     & 0.751                                     & 0.725                                     & 0.724                                     & 0.760                                      \\
Min-60\%++              & 0.696                                     & 0.724                                     & 0.570                                     & 0.606                                     & 0.591                                     & 0.525                                     & 0.619                                     &  & 0.774                                     & 0.786                                     & 0.788                                     & 0.752                                     & 0.727                                     & 0.725                                     & 0.759                                      \\
Min-70\%++              & {\cellcolor[rgb]{0.784,0.784,0.784}}0.698 & 0.720                                     & 0.567                                     & 0.605                                     & 0.590                                     & 0.525                                     & 0.617                                     &  & 0.774                                     & 0.785                                     & 0.787                                     & 0.753                                     & 0.731                                     & 0.726                                     & 0.759                                      \\
Min-80\%++              & 0.697                                     & 0.718                                     & 0.562                                     & 0.605                                     & 0.592                                     & 0.525                                     & 0.616                                     &  & 0.771                                     & 0.785                                     & 0.785                                     & 0.754                                     & 0.736                                     & 0.729                                     & 0.760                                      \\
Min-90\%++              & 0.697                                     & 0.716                                     & 0.558                                     & 0.604                                     & {\cellcolor[rgb]{0.784,0.784,0.784}}0.593 & 0.525                                     & 0.615                                     &  & 0.769                                     & 0.788                                     & 0.782                                     & 0.758                                     & 0.739                                     & 0.731                                     & 0.761                                      \\
Min-100\%++              & 0.695                                     & 0.713                                     & 0.551                                     & 0.602                                     & 0.591                                     & 0.525                                     & 0.613                                     &  & 0.766                                     & 0.787                                     & 0.770                                     & 0.750                                     & 0.732                                     & 0.726                                     & 0.755                                      \\
\hline
\end{tabular}
}
\end{table*}

\begin{table*}
\centering

\caption{The accuracy of original and privacy-preserving LLMs across all 57 tasks of MMLU}\label{tab:mmlu}
\resizebox{0.8\linewidth}{!}{%

\begin{tabular}{llll}
Task                                    & Accuracy (w/ Defender) & Accuracy (w/o Defender) & supercategory    \\ 
\hline
abstract\_algebra                       & 0.497                  & 0.491                   & STEM             \\
anatomy                                 & 0.558                  & 0.497                   & STEM             \\
astronomy                               & 0.515                  & 0.485                   & STEM             \\
business\_ethics                        & 0.552                  & 0.503                   & Other            \\
clinical\_knowledge                     & 0.582                  & 0.479                   & Other            \\
college\_biology                        & 0.539                  & 0.485                   & STEM             \\
college\_chemistry                      & 0.491                  & 0.533                   & STEM             \\
college\_computer\_science              & 0.545                  & 0.503                   & STEM             \\
college\_mathematics                    & 0.588                  & 0.461                   & STEM             \\
college\_medicine                       & 0.533                  & 0.509                   & Other            \\
college\_physics                        & 0.485                  & 0.527                   & STEM             \\
computer\_security                      & 0.521                  & 0.545                   & STEM             \\
conceptual\_physics                     & 0.515                  & 0.442                   & STEM             \\
econometrics                            & 0.497                  & 0.515                   & Social Sciences  \\
electrical\_engineering                 & 0.473                  & 0.503                   & STEM             \\
elementary\_mathematics                 & 0.533                  & 0.497                   & STEM             \\
formal\_logic                           & 0.509                  & 0.473                   & Humanities       \\
global\_facts                           & 0.515                  & 0.521                   & Other            \\
high\_school\_biology                   & 0.558                  & 0.527                   & STEM             \\
high\_school\_chemistry                 & 0.497                  & 0.497                   & STEM             \\
high\_school\_computer\_science         & 0.521                  & 0.533                   & STEM             \\
high\_school\_european\_history         & 0.552                  & 0.473                   & Humanities       \\
high\_school\_geography                 & 0.515                  & 0.533                   & Social Sciences  \\
high\_school\_government\_and\_politics & 0.521                  & 0.545                   & Social Sciences  \\
high\_school\_macroeconomics            & 0.485                  & 0.545                   & Social Sciences  \\
high\_school\_mathematics               & 0.552                  & 0.503                   & STEM             \\
high\_school\_microeconomics            & 0.503                  & 0.497                   & Social Sciences  \\
high\_school\_physics                   & 0.509                  & 0.521                   & STEM             \\
high\_school\_psychology                & 0.479                  & 0.570                   & Social Sciences  \\
high\_school\_statistics                & 0.509                  & 0.533                   & STEM             \\
high\_school\_us\_history               & 0.491                  & 0.491                   & Humanities       \\
high\_school\_world\_history            & 0.473                  & 0.497                   & Humanities       \\
human\_aging                            & 0.527                  & 0.473                   & Other            \\
human\_sexuality                        & 0.509                  & 0.521                   & Social Sciences  \\
international\_law                      & 0.509                  & 0.497                   & Humanities       \\
jurisprudence                           & 0.515                  & 0.521                   & Humanities       \\
logical\_fallacies                      & 0.533                  & 0.515                   & Humanities       \\
machine\_learning                       & 0.558                  & 0.461                   & STEM             \\
management                              & 0.521                  & 0.527                   & Other            \\
marketing                               & 0.545                  & 0.491                   & Other            \\
medical\_genetics                       & 0.473                  & 0.527                   & Other            \\
miscellaneous                           & 0.539                  & 0.539                   & Other            \\
moral\_disputes                         & 0.497                  & 0.509                   & Humanities       \\
moral\_scenarios                        & 0.521                  & 0.509                   & Humanities       \\
nutrition                               & 0.533                  & 0.539                   & Other            \\
philosophy                              & 0.509                  & 0.491                   & Humanities       \\
prehistory                              & 0.509                  & 0.485                   & Humanities       \\
professional\_accounting                & 0.521                  & 0.533                   & Other            \\
professional\_law                       & 0.479                  & 0.479                   & Humanities       \\
professional\_medicine                  & 0.539                  & 0.533                   & Other            \\
professional\_psychology                & 0.533                  & 0.479                   & Social Sciences  \\
public\_relations                       & 0.515                  & 0.545                   & Social Sciences  \\
security\_studies                       & 0.491                  & 0.497                   & Social Sciences  \\
sociology                               & 0.539                  & 0.521                   & Social Sciences  \\
us\_foreign\_policy                     & 0.503                  & 0.521                   & Social Sciences  \\
virology                                & 0.527                  & 0.515                   & Other            \\
world\_religions                        & 0.467                  & 0.576                   & Humanities       \\
\hline
\end{tabular}

}
\end{table*}

\newpage

\subsection{Detailed Experimental Settings}\label{par:detailed_settings}
\subsubsection{Target models}
WIKIMIA and WIKIMIA-24 are both adaptive to a wide range of LLMs because Wikipedia articles are typically included in the training corpus of LLMs~\cite{shi2023detecting}. 
However, the WIKIMIA benchmark is only feasible for evaluating the LLMs that were released before 2023, while our constructed WIKIMIA-24 benchmark can evaluate the LLMs that were released before March 2024. Thus, we consider the following LLMs with both aligned and unaligned versions: Pythia~\cite{biderman2023pythia}, Falcon~\cite{falcon40b}, LLaMA~\cite{touvron2023llamaa}, LLaMA-2~\cite{touvron2023llama}, Mistral~\cite{jiang2023mistral}, Gemma~\cite{team2024gemma}, and only Pythia, Falcon and LLaMA are evaluated on WIKIMIA. All models are loaded from Huggingface with the transformers package~\cite{wolf-etal-2020-transformers}, the detailed information of these target models including the URL can be found in Table~\ref{tab:target_llm}.

\subsubsection{Baselines}
We employ seven representative MIA methods designed for LLMs, which were evaluated or proposed in previous works~\cite{shi2023detecting, zhang2024mink}. 
These baselines include three metric-based and four calibration-based MIA methods. 
The metric-based methods are based solely on a certain metric of the target text evaluated in the target model: 1) \textbf{PPL}~\cite{yeom2018privacy}: a basic method that used the PPL or the loss as the MIA metric 2) \textbf{Min-k\%}~\cite{shi2023detecting}: evaluate the token-level probability and adopt the average over the $k\%$ minimum token probability as the MIA metric 3) \textbf{Min-K\%++}~\cite{zhang2024mink}: an enhanced version of Min-k\% that detects pre-training recrods by auditing whether the record forms a relatively high probability under the conditional categorical distribution of the predicted next token.
The calibration-based methods use a certain reference metric as a calibration item for the original metric: 4) \textbf{Zlib}~\cite{carlini2021extracting}: using the zlib compression entropy as the reference metric. 5) \textbf{Lowercase}~\cite{carlini2021extracting}: using the PPL of the lowercased text as the reference metric. 6) \textbf{Neighbor}~\cite{mattern2023membership}: the reference metric is designed as the PPL of the neighboring texts, which are generated by replacing partial tokens of the target text with a mask-filling language model. 7) \textbf{Neighbor}~\cite{mattern2023membership}: using the PPL measured on a smaller version of the target LLM as the reference metric. The attack strategy of each baseline is identical on both aligned and unaligned LLMs.


\subsubsection{Implementation Details}
All experiments are compiled and tested on a Linux server (CPU: AMD EPYC-7763, GPU: NVIDIA GeForce RTX 3090), we spent around 21 days to accomplish all the experiments. We adopted 160 samples (80 member and 80 non-member samples) for tuning both aligned and unaliged LLMs with a batch size of 16 and a learning rate of 0.0005. All LLMs are tuning for 20 epoches. The pending text length is set to 128 tokens. We adopt the AdamW optimizer for fine-tuning. For the hybrid loss of aligned LLMs, we set the weights to $\alpha=1$, $\beta=1$, $\gamma=1$. For the contrastive loss designed for unaligned LLMs, we set the temperature $\tau=10$.

\subsection{Evaluation Results on WikiMIA Dataset}\label{par:wikimia_performance}

The evaluation results of MIA-Tuner and all baselines can be found in Table~\ref{tab:wikimia_performance}.

\subsection{Evaluation Some Baselines with Different Parameters}\label{par:differentk}
We evaluation Min-K\% and  Min-K\%++ in WikiMIA-24 across all target models, the resluts are summarized in Table~\ref{tab:differentk}.

\subsection{Raw Evaluation Results on MMLU}\label{par:mmlu}
We provide the accuracy of original and privacy-preserving LLMs across all 57 tasks of MMLU in Table~\ref{tab:mmlu}.

\clearpage

\end{document}